\documentclass[letterpaper, 10 pt, journal, twoside]{IEEEtran}      

\IEEEoverridecommandlockouts                              

\usepackage[pdftex]{graphicx}
\usepackage{tabularx}
\usepackage{multirow}
\usepackage{array}
\usepackage{booktabs}
\usepackage{amsmath}
\usepackage{amssymb}
\usepackage{amsfonts}
\usepackage[ruled, vlined, linesnumbered]{algorithm2e}

\usepackage{algorithmic}

\usepackage{balance}

\usepackage{graphicx, subcaption}

\usepackage{url}
\usepackage[colorlinks=true,urlcolor=blue]{hyperref} 
\hypersetup{
    linkcolor = {black},
    citecolor = {black}
}

\usepackage{xcolor}
\DeclareMathOperator*{\argmax}{arg\,max}

\usepackage[font=small]{caption}

\usepackage{balance}

\usepackage{todonotes}

\title{A Shadowcasting-Based Next-Best-View Planner for Autonomous 3D Exploration}

\author{Ana Batinovic, Antun Ivanovic, Tamara Petrovic and Stjepan Bogdan
\thanks{Authors are with the University of Zagreb, Faculty of Electrical Engineering  and Computing, LARICS Laboratory for Robotics and Intelligent Control Systems, Unska 3, 10000 Zagreb, Croatia; {\tt\small (ana.batinovic, antun.ivanovic, tamara.petrovic, stjepan.bogdan)@fer.hr}}}

\begin{document}

\maketitle

\begin{abstract}
In this paper, we address the problem of autonomous exploration of unknown environments with an aerial robot equipped with a sensory set that produces large point clouds, such as LiDARs.
The main goal is to gradually explore an area while planning paths and calculating information gain in short computation time, suitable for implementation on an on-board computer. To this end, we present a planner that randomly samples viewpoints in the environment map. It relies on a novel and efficient gain calculation based on the Recursive Shadowcasting algorithm. To determine the Next-Best-View (NBV), our planner uses a cuboid-based evaluation method that results in an enviably short computation time. To reduce the overall exploration time, we also use a dead end resolving strategy that allows us to quickly recover from dead ends in a challenging environment. 
Comparative experiments in simulation have shown that our approach outperforms the current state-of-the-art in terms of computational efficiency and total exploration time. The video of our approach can be found at \href{https://www.youtube.com/playlist?list=PLC0C6uwoEQ8ZDhny1VdmFXLeTQOSBibQl}{https://www.youtube.com/playlist?list=PLC0C6uwoEQ8ZDhny1V dmFXLeTQOSBibQl}.
\end{abstract}

\IEEEpeerreviewmaketitle

\section{Introduction}
\label{sec:introdiction}

In this paper, we study an autonomous exploration and mapping of a completely unknown 3D environment. We propose a novel method that improves upon the state-of-the-art Receding Horizon Next-Best-View planing (RH-NBVP) \cite{Bircher2016}, which uses a sampling-based approach to select the next best viewpoint \cite{GonzalezBanos2002}. This planner is used in combination with Rapidly-exploring Random Trees (RRT) \cite{RRT1}, \cite{RRT2} to generate traversable paths. For each node in the RRT path, the information gain is calculated as a volume of the unmapped space that would be observed by robot sensors when the robot is positioned in the target node. A common algorithm used for the information gain estimation is the Raycasting algorithm (RC) \cite{Elvins1992}, and its results are then weighted by the cost of the robot travelling to the node. The best RRT path is then determined and the first edge is traversed before running a new iteration of the path planner.

The main drawbacks of the RH-NBVP are the significant computation time required to compute the information gain using the Raycasting algorithm and the high probability of ending up in a dead end state during the exploration. To overcome these issues, we propose a new strategy based on a Recursive Shadowcasting (RSC) algorithm,  proposed in \cite{Bergstrom2001}. Since the RSC allows a much faster computation, we can estimate the information gain not only for each RRT node, but also for each RRT edge. We propose a cuboid-based evaluation for each RRT edge to obtain a more complete information about the unknown space to be discovered. We select the best RRT path and execute a trajectory through the RRT nodes of the best path.

We extend the RH-NBVP to deal with sensors that produce large point clouds with each scan, such as LiDARs. Since we use a large point cloud, the RC in the information gain calculation process increases the computation time \cite{Bircher2016}, \cite{Selin2019}, \cite{Respall2021}. On the other hand, using LIDARs in combination with the RSC results in a significant computation time reduction during planning iterations.
 
The RRT has its root in the current position of the robot and is recomputed in each iteration. In large and narrow environments, as the explored area increases, the RRT algorithm might end up stuck in a dead end characterized by a significant increase in the distance to the node with a non-zero information gain and in the time required to sample valid RRT nodes. To address this drawback, motivated by the previous work on history-aware approaches to the 3D exploration (\cite{Witting2018}, \cite{Selin2019}, \cite{Respall2021}), we developed a method to resolve such states by tracking previously visited RRT nodes.

\begin{figure}[t!]
	\centering
	\includegraphics[width=1\columnwidth]{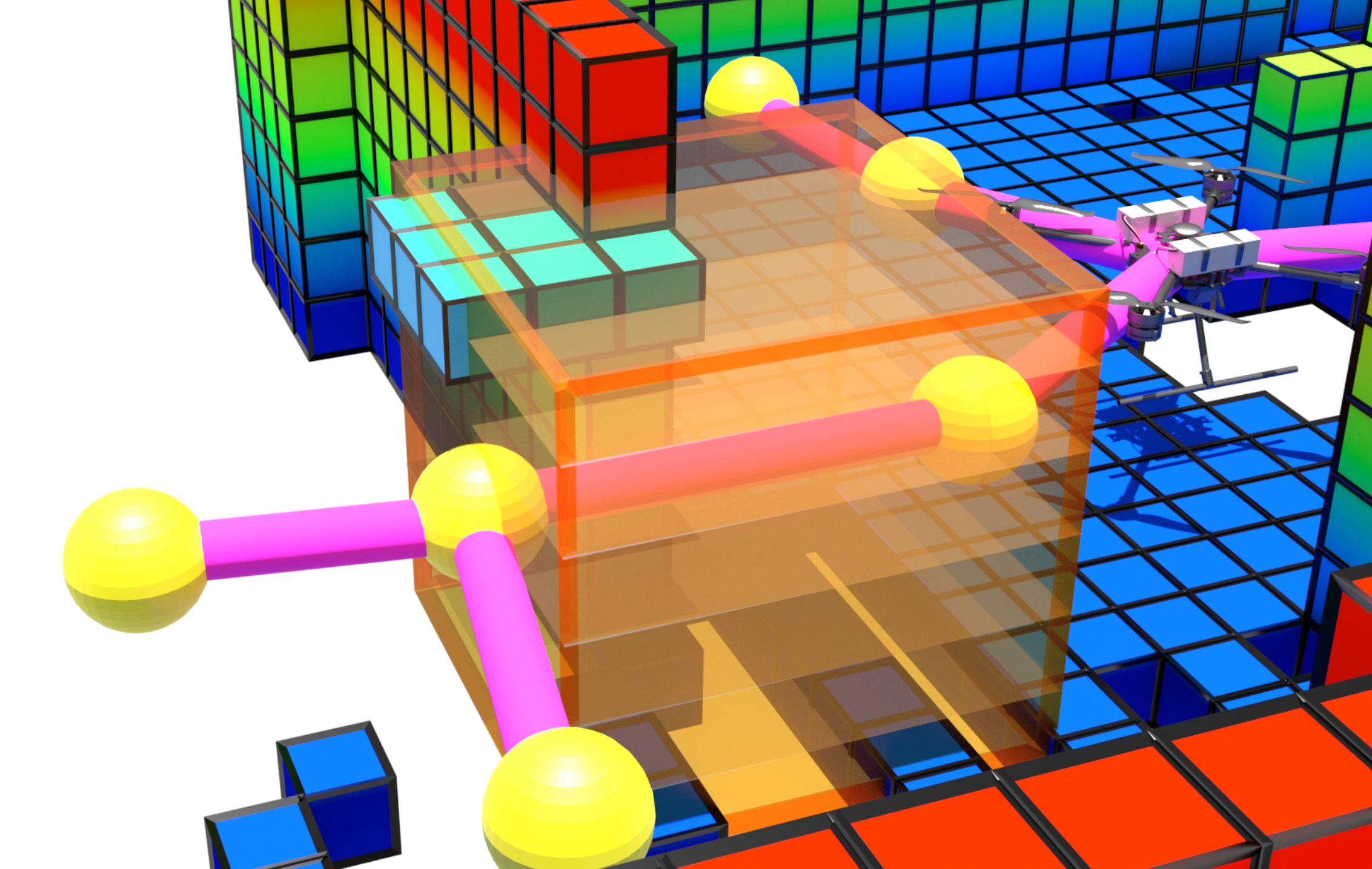}
	\caption{An illustration of the proposed Recursive Shadowcasting algorithm in the OctoMap. The algorithm is performed inside the cuboid centered at the path edge, on each 2D slice (planes inside the cuboid). The results of the RSC are shown on the first OctoMap slice, where the cyan voxels represents unknown voxels while the grey voxel is not taken into account for the information gain calculation.}
	\label{fig:shadowcasting-cuboid}
\end{figure}

We compare our method with the state-of-the-art methods in the simulation. The results show that in all cases our method achieves the complete exploration faster with an enviably low computation time. The contributions of this paper are summarized as follows:
\begin{itemize}
    \item Time-efficient information gain estimation based on the Recursive Shadowcasting algorithm.
    \item Cuboid-based estimation of information gain on each RRT edge.
    \item A history tracking method for resolving dead end states.
\end{itemize}

To validate our contributions, we performed an extensive simulation analysis and comparison with the state-of-the-art approaches. Furthermore, to encourage the reproduction of our results and facilitate more thorough future comparisons in the exploration field of research, the source code, data sets of simulations and experiments carried out for preparation of this paper are available at \cite{open-source}.

In Section \ref{sec:related} we give an overview of the state-of-the-art of 3D exploration methods and position our work in relation to them. Section \ref{sec:proposed} is the core of the paper and contains details of the proposed planner. Results of simulations performed with a Unmanned Aerial Vehicle (UAV) and their analysis are presented in Section \ref{sec:simulation}. The paper ends with a conclusion in Section \ref{sec:conclusion}.
\vspace{-0.35cm}

\section{Related work}
\label{sec:related}

Autonomous exploration and mapping is one of the fundamental tasks of robotics. Typical exploration approaches can be roughly classified into frontier-based, sampling-based, and hybrid strategies.  

Characteristic of frontier-based approaches is exploration by approaching a selected point on the frontier between the explored and unexplored environments. This idea was first introduced by Yamauchi in \cite{Yamauchi1997} and subsequently evaluated in more detail in \cite{Julia2012}. In each iteration, the next best goal is a frontier point closest to the robot. Similarly, in \cite{Cieslewski2017}, the next best goal is the frontier that minimizes the velocity change to maintain a consistently high flight speed. It is shown that this approach outperforms the closest frontier method \cite{Yamauchi1997}. 
Frontier-based exploration approaches for 3D environments are also researched in \cite{Zhu2015}, \cite{Mannucci2017}, \cite{Batinovic-RAL-2021}, \cite{Dai2020}, \cite{Faria2019}, \cite{Zhou2021}.

Sampling-based approaches aim to determine a (minimal) sequence of robot (sensor) viewpoints to visit in the environment until the entire space is explored. Potential viewpoints are typically sampled, e.g., near the frontier or randomly. Then these viewpoints are evaluated for the potential information gain and the next best viewpoint is assigned. One of the first sampling-based methods is presented in \cite{GonzalezBanos2002} and then extended in \cite{Baiming2018}, \cite{Bircher2016}, \cite{Joho2007}. In \cite{Bircher2016}, authors proposed the NBV planner (RH-NBVP), which uses an RRT-based search \cite{RRT1}, \cite{RRT2} to guide a UAV into the unexplored area. While the method showed good scaling properties and performance in a local exploration, it is not resilient to dead ends, resulting in a poor global scene coverage and thus, a high overall exploration time, as shown in \cite{Cieslewski2017}, \cite{Deng2020}, \cite{Dai2020}, and in our previous work \cite{Batinovic-RAL-2021}. To address the drawbacks of the RH-NBVP, Witting et al. \cite{Witting2018} introduced several modifications: memorizing previously visited locations; local gain optimization; and trajectory optimization, resulting in faster exploration. In \cite{Schmid2020}, authors improve the efficiency of RH-NBVP by continuously growing a single tree and only sporadically querying feasible paths.  
 
Hybrid strategies combine the advantages of both frontier-based and sampling-based approaches. Selin et al. \cite{Selin2019} successfully combine RH-NBVP with conventional frontier reasoning to compensate for a poor performance in the global exploration. In other words, \cite{Selin2019} plans global paths towards frontiers and samples paths locally. Meng et al. \cite{Meng2017} samples viewpoints around frontiers and finds the global shortest tour passing through them. Similarly, Respall et al. \cite{Respall2021} samples viewpoints in the vicinity of a point of interest near a frontier and additionally memorizes nodes that indicate regions of interest in a history graph to reduce the gain calculation time. Song et al. \cite{Song2017} generate inspection paths that completely cover the frontier using a sampling-based algorithm. 

In summary, we find that RH-NBVP is a very promising method for general 3D exploration. Therefore, we build our method on the RH-NBVP and introduce several modifications that lead to shorter computation times and faster exploration.

\section{Proposed approach}
\label{sec:proposed}

\subsection{System overview}
The main goal of our approach is to explore a bounded and previously unknown 3D space $V \subset \mathbb{R}^{3}$. 
As a basis for our approach we use an OctoMap, a hierarchical volumetric 3D representation of the environment \cite{Hornung2013}. Each cube of the OctoMap is denoted as a voxel (cell), which can be \textit{free}, \textit{occupied} or \textit{unknown}. Free voxels form the free space $V_{free} \subset V$, occupied voxels form the occupied space $V_{occ} \subset V$ and unknown voxels form the unknown space $V_{un} \subset V$. Initially, the entire bounded space is unknown, $V \equiv V_{un}$, and the unknown space decreases as the exploration advances. The entire space is a union of the three subspaces $V \equiv V_{free} \cup V_{occ} \cup V_{un}$. The goal of the exploration process is to explore the environment as soon as possible.

\begin{figure}[t!]
	\centering
	\includegraphics[width=1\columnwidth]{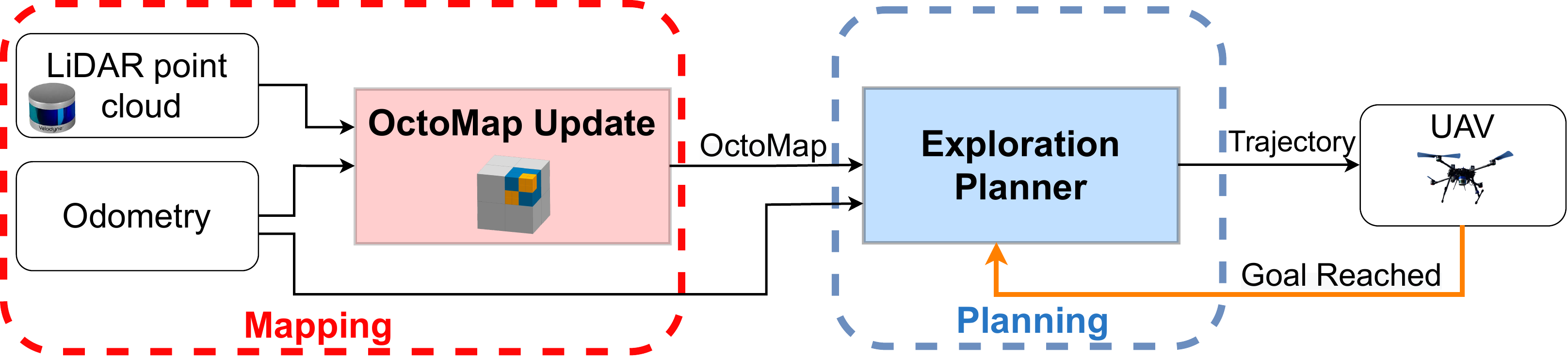}
	\caption{Overall schematic diagram of the 3D exploration. The LiDAR point cloud and odometry data represent inputs to the OctoMap module. The exploration planner module (highlighted in blue) generates a trajectory to the selected target towards which the robot navigates. \vspace{-0.25cm}}
	\label{fig:system}
\end{figure}

Our proposed approach is a sampling-based exploration where the goal is to increase the overall knowledge of the environment by directing the robot in a way that reduces the overall exploration time. An overview of the proposed system is given in Fig. \ref{fig:system}. The OctoMap module requires a suitable sensing system, such as a laser scanner or a camera, to create a 3D map. In our case, a LiDAR point cloud is used to generate an OctoMap, which is used for both exploration and collision-free navigation.

The exploration method is based on a novel information gain computation algorithm that ensures a fast exploration of the environment. We use RSC algorithm to calculate the \textit{information gain} for each RRT edge, evaluating the whole path rather than the subsequent point only, and navigate the robot towards the best path. Our approach leads to an efficient global exploration of the environment. 

\subsection{UAV and Sensor Models}

In this work, the exploration is performed with a UAV that has no prior knowledge of the environment. Although the concepts are explained with the UAV in mind, the same approach is applicable on other autonomous robots equipped with LiDARs or other sensors that can be used to build an OctoMap. 

The UAV is represented with a state vector $\mathbf{x} = \begin{bmatrix} \mathbf{p}^T & \psi \end{bmatrix}^T \in \mathbb{R}^4$ that consists of the position $\mathbf{p} = \begin{bmatrix} x & y & z \end{bmatrix}^T \in \mathbb{R}^{3}$ and the yaw rotation angle around z axis $\psi \in [-\pi, \pi)$. Furthermore, the algorithm assumes a maximum linear velocity $\mathbf{v}_{max} \in \mathbb{R}^{3}$ and a maximum angular velocity around z axis $\dot{\psi}_{max}$. For collision checking, it is considered that the UAV is inside a prism centered at $\mathbf{p}$, with adequate length, width and height $l$, $w$, $h$.
The algorithm relies on a maximum range of the sensor $R_{max} \in \mathbb{R}$ with horizontal and vertical Field of View (FOV) in range, $\alpha_{h}$, $\alpha_{v}$ $\in (0^{\circ}, 360^{\circ}]$, respectively. This allows our algorithm to work with point-cloud-producing sensors with various FOV, such as camera with limited FOV and LiDARs with limited  $\alpha_{v}$.

\subsection{Overview of RH-NBVP and Raycasting algorithms}

The RH-NBVP samples nodes from the position of the robot using the RRT algorithm. For each new node, the expected information gain is calculated as the sum of the unknown volume in the sampled camera FOV, exponentially weighted by distance from the node to the current position of the robot. The total gain of a node is the sum of all gains along the RRT path to that node. The growth of the tree is limited with a predetermined number of nodes. When the limit is reached, the node with the highest total information gain defines the best path and the robot executes only the first edge of the path. The described procedure is then repeated. The exploration is considered complete when the best node information gain is below a threshold $g_{zero}$ and the tree reaches the maximum number iterations.

Information gain for each node is calculated using a Raycasting algorithm which traces the path of a series of rays originating from a given node. The density and range of rays define the area to be examined and are specified in advance. When one of these rays hits an obstacle (e.g., a wall), all voxels that the ray previously touched are considered as free voxels. Otherwise, the voxels are considered as free or unknown, depending on the current state of the OctoMap. This results in knowing the number of free and unknown voxels in a predefined area, in each direction from a specific position. Based on this information, a robot can take appropriate actions to move to an unknown area to reduce the total exploration time. 

In general, all algorithms that directly cast rays into the map, cast more rays than necessary because they cast a fixed number of rays regardless of the design of the environment \cite{Evan2021}. It is shown in \cite{Evan2019-diss} that the computation time of Raycasting algorithm increases as the predefined area increases. This is because the number of rays depends on the predefined area and is not affected by the occupied voxels (obstacles). The problem of computational effort required to calculate the information gain becomes even more apparent when using sensors that produce large point clouds, such as LiDARs.

Moreover, in RH-NBVP, the robot moves to the first node of the best path before performing another planning iteration. This may cause the robot to move back and forth in a small area, changing the best path in each iteration. As the size of the explored area increases, the RRT can easily reach the maximum number of iteration and result in uncovered regions. Moreover, if the distance to the next node is large, the RRT sampling time increases significantly. This limitation is usually noticeable in narrow and large environments.

\subsection{Recursive Shadowcasting Algorithm Overview}
\label{subsec:shadowcasting}



\begin{figure}[t!]
	\centering
	\includegraphics[width=0.8\columnwidth]{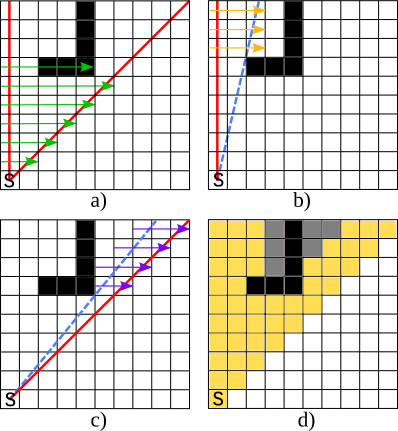}
	\caption{Steps of the RSC on a single octant. \vspace{-0.25cm}}
	\label{fig:shadow_casting_steps}
\end{figure}

Recursive Shadowcasting was first used in computer games to calculate a FOV from a top-down perspective, where the FOV is defined as a set of locations visible from a specific position in a computer game scene \cite{Evan2021}. The original RSC, proposed in \cite{Bergstrom2001}, considers a 2D FOV grid and initially sets all grid cells to not visible. Then the grid is divided into eight octants centred on the FOV source ($\mathbf{S}$) and the cells within each octant are traversed \cite{Evan2021}. This traversal occurs within each octant by rows or columns in ascending order of distance from the FOV source. Fig. \ref{fig:shadow_casting_steps} shows the steps of the RSC on an octant. When a cell is traversed, its visibility state is set to visible. However, when an occupied cell (the black cell) is encountered, an octant is recursively split into smaller regions (Fig. \ref{fig:shadow_casting_steps} b) and c)), which are bounded by rays cast from the FOV source to the corners of the occupied cell (blue dashed rays). The cell traversals are then continued within each smaller region. As marked in Fig. \ref{fig:shadow_casting_steps} a) with green arrows, the algorithm first processes rows one through five without encountering any occupied cell. In line six, three occupied cells are encountered, splitting the free region in two and causing the algorithm to call itself recursively. The recursive call then processes the free region on the left (Fig. \ref{fig:shadow_casting_steps} b)), while the main iteration of the algorithm continues processing the free region on the right. Note that even if a ray only grazes the edge of a cell, that entire cell is set to visible.

The result of the RSC on an octant in 2D is shown in Fig. \ref{fig:shadow_casting_steps} d), where occupied cells are shown in black, visible cells in yellow and invisible cells in grey. Similarly, the main goal of the RSC in the information gain calculation is to find unknown voxels of the OctoMap among the visible cells.
 
Similar to the FOV grid in 2D computer games, a 3D OctoMap used in this paper is divided into cube-shaped voxels, allowing us to take advantage the RSC in a 2.5D environment for the information gain calculation. To the best of our knowledge, this is the first time that the RSC algorithm is used in the exploration of the environment. In the next section we show how the RSC is used for the evaluation of RRT paths.

\subsection{Cuboid-Based Path Evaluation}
\label{subsec:cuboid}

The RRT algorithm samples nodes $\mathbf{n} = \begin{bmatrix} x & y & z \end{bmatrix}^T \in \mathbb{R}^{3}$. A collision-free RRT path is denoted as $\mu \in M$, where $M$ denotes the set of all RRT paths. Let $\mu_j$, $j \in (1,2,...,N)$ be the path edge between nodes $\mathbf{n}_{k-1}$ and $\mathbf{n}_k$, where $k \in (1,2,...,N+1)$ and $N$ is the number of nodes. For path edge $\mu_j$,  we define information gain $I(\mu_j) \in \mathbb{R}$ as a measure of an unexplored region of the environment that is potentially visible from the center $\mathbf{c}_k$ of this path edge. 

To determine the information gain $I(\mu_j)$ using the RSC, we first place a cuboid around the edge $\mu_j$. The cuboid center point is $\mathbf{c}_k$, while the cuboid length is $l_c = \lVert \mathbf{n}_k - \mathbf{n}_{k-1}\rVert$. The cuboid width $w_c$ and height $h_c$ are defined by the parameter $I_{range}$, which depends on the used sensor range and the environment size. 
The point $\mathbf{c}_k$ is the FOV grid source inside which the RSC is performed, while the cuboid determines the borders of each FOV grid. The FOV grid is obtained as a 2D slice of the OctoMap at point $\mathbf{c}_k$ and the RSC is performed 360$^\circ$ with a horizontal step size $r$.  
An illustration of the cuboid centered at the path edge is shown in Fig. \ref{fig:shadowcasting-cuboid}. We simplified the illustration showing the performance of the RSC on the first 2D OctoMap slice. As described, the algorithm is performed on each slice and on each path edge.

Note that the maximum cuboid length $l_{max}$ is predefined according to the size of the environment, because calculating the information gain on a large path edge using only one center, that is, one FOV grid for RSC, may result in missing information and poor information gain calculation. In other words, when $l_c > l_{max}$ we add intermediate FOV sources to cover the path edge and to achieve $l_c < l_{max}$.
 
To form the information gain of the node $\mathbf{n}_k$, edge information gain $I(\mu_j)$ is weighted with the negative exponential of the cost to travel along the path up to $\mathbf{n}_k$, similar to the one proposed in \cite{GonzalezBanos2002} and used in \cite{Bircher2016}:
\begin{equation}
    I(\mathbf{n}_k) = I(\mathbf{n}_{k-1}) + \frac{I(\mu_j)}{e^{\lambda L(\mathbf{n}_k, \mathbf{n}_{k-1})}},
\label{eq:node_gain}
\end{equation}
where $\lambda$ is a positive constant, $L(\mathbf{n}_k, \mathbf{n}_{k-1})$ is Euclidean distance between nodes $\mathbf{n}_k$ and $\mathbf{n}_{k-1}$. The constant $\lambda$ weighs the importance of the robot motion cost against the expected information gain. A small $\lambda$ gives the priority to the information gain, while $\lambda \rightarrow \infty $ means that the motion is so expensive that the shortest path is selected.
The value of $\lambda$ is experimentally determined.


As for the complexity of algorithms in the information gain calculation, performing a single RC with $n$ horizontal and $m$ vertical rays with the resolution $r$ of the map scales with $\mathcal{O}(\frac{1}{r^4})$ in \cite{Bircher2016}, $\mathcal{O}(nm/r)$ in \cite{Selin2019}, while our approach with the RSC scales with $\mathcal{O}(\frac{mn\log{n}}{r})$. 
By using the proposed algorithm, the high calculation effort required by the RC is avoided. The main reason for the calculation effort reduction is the property of the RSC that ensures each voxel is visited only once.

As presented in \cite{Evan2019-diss}, \cite{Jice2009}, RSC has significantly better performance among existing FOV algorithms. However, \cite{Evan2021} showed the drawback of RSC algorithm when the grid size increases to tens of thousands of cells. This is because it performs a relatively large number of operations per cell. Nevertheless, we find the RSC suitable for our exploration environments, sensor specifications and OctoMap size. 


\subsection{Exploration of the Best Path}
According to Eq. \ref{eq:node_gain}, path information gain $I(\mu)$ is equal to information gain of the path's last node and presents the volume of the unknown space that is covered along the path, combined with the cost of going there.

In each iteration, our goal is to find the best path $\mu_{bp}$, which maximizes the information gain  $I(\mu)$:

\begin{equation}
    \mu_{bp} = \argmax_{\mathbf{\mu} \in M} I(\mu).
\end{equation} 

As soon as the best path $\mu_{bp}$  is selected, we address the yaw angle along that path. In RH-NBVP the yaw angle is randomly sampled during the exploration, which limits the sample efficiency of the exploration. This limitation is briefly addressed in \cite{Witting2018}, \cite{Respall2021}, and is not the scope of this paper. Since we use a LiDAR sensor with horizontal FOV $\alpha_{h} = 360 ^\circ$, which is attached to the UAV with some pitch angle, our strategy is to align the yaw angle towards the next point on the path: 
\begin{equation}
     \label{eqn:yaw_alignment}
     \psi_{k} = \arctan{\dfrac{y_{k}-y_{k-1}}{x_{k}-x_{k-1}}}, k \in \left\{2 \dots n \right\},
 \end{equation}
\noindent where n is the total number of points in the planned path. Note that for the first waypoint, $k=1$, we use the current UAV orientation.

\begin{figure}[t!]
	\centering
	\includegraphics[width=0.99\columnwidth]{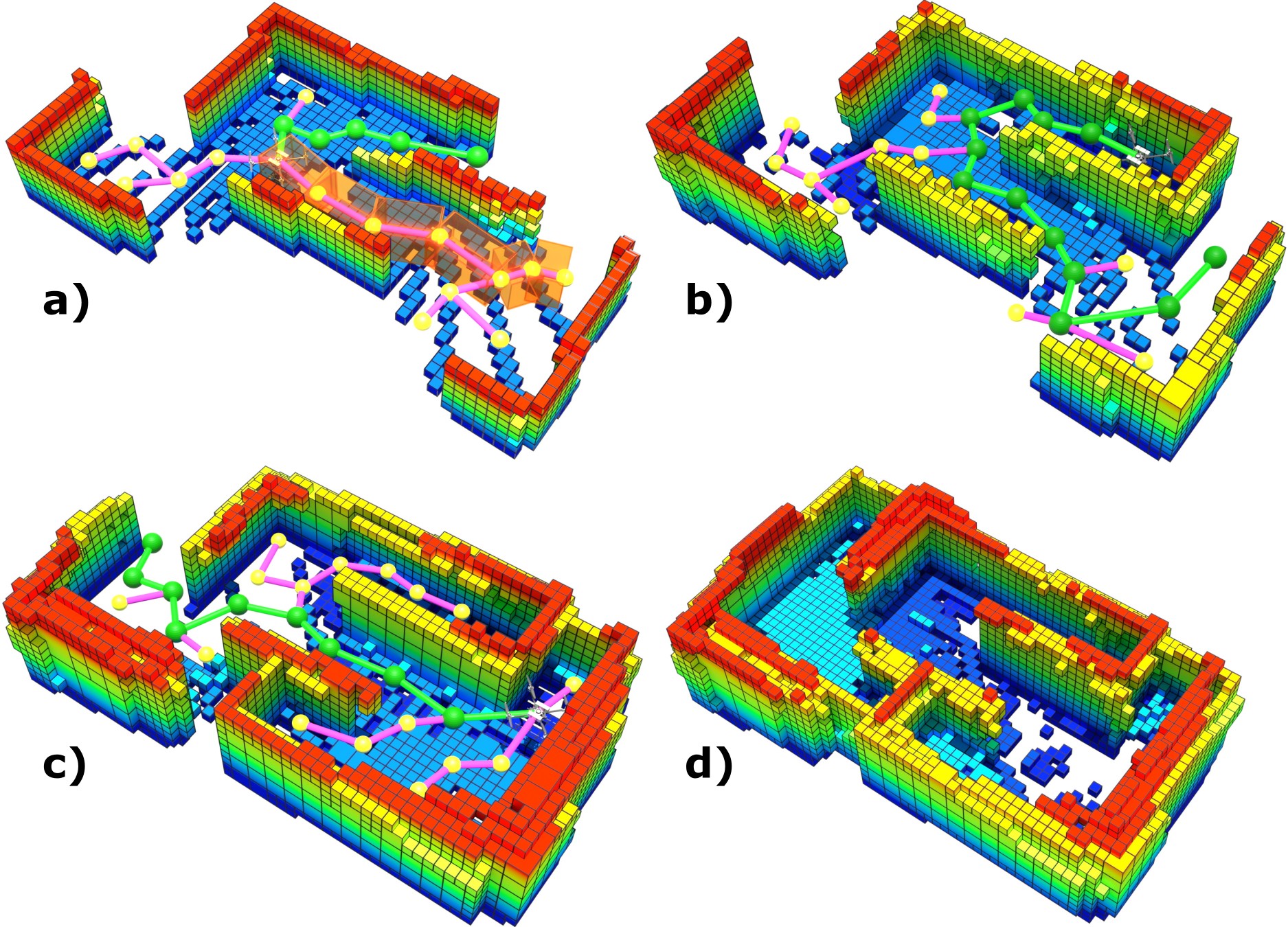}
	\caption{An illustration of the exploration process. Green path with green waypoints is the best path. Purple paths with yellow waypoints are other paths of the RRT. a) The initial tree with selected path leading towards upper right portion of the environment. Orange cuboids along the right path of the RRT illustrate the volume where the information gain is computed. b) The second iteration with tree leading towards the right of the environment. c) The third iteration leading towards the left portion of the environment. d) Exploration finished and the final map of the environment obtained after third iteration. \vspace{-0.5cm}}
	\label{fig:octomap_exploration}
\end{figure}

After the path has been augmented with the yaw angle, it is forwarded to the the trajectory planner. Within this paper, we use the Time Optimal Path Parametrization by Reachability Analysis (TOPP-RA) algorithm developed in \cite{toppra}. Apart from the waypoints, inputs for the TOPP-RA are also velocity and acceleration constraints, which are maximally set to the UAV physical limitations. The planned trajectory is then executed by the UAV, and a new cycle for determining the best path is started after the UAV stops. The exploration process is performed until the entire environment is explored, yielding the environment map. The described process is depicted on Fig. \ref{fig:octomap_exploration}.

\subsection{Dead End Resolving Strategy}

One of the drawbacks of the RH-NBVP algorithm is the dead end state. It especially occurs in large and narrow environments where the RRT algorithm might end up stuck in a dead end, trying to grow the tree up to the node with non-zero information gain. This results in both higher computation and total exploration time.
Inspired by the idea from \cite{Witting2018}, we propose a different approach to resolve a dead end state and mitigate the effects of computational requirements needed to continue the exploration. 

Dead end is a state in which a robot is unable to find a feasible path to an unexplored region, even though such a region exist. This is often the case when such regions are distant and thus, have very low information gains, due to parameter $\lambda$ described in \ref{subsec:cuboid}, which is a trade-off between the information gain and the distance. 

When the robot is in a dead end, the main idea in order to resolve this state is to return the robot to a previously-visited node that has the highest information gain at the moment. In that manner we try to maximize the probability of finding a feasible path to the unexplored regions in a single iteration. To do so, in each iteration all previously visited nodes and their information gains are stored. We denote the node with the highest information gain $\mathbf{n}_{bn}$, and call it the best node. Return to the best node is described in Algorithm \ref{alg:recovery}. 

The node $\mathbf{n}_{bn}$ represents the desired node to return to. Starting from the $\mathbf{n}_{bn}$ up to the current node  $\mathbf{n}_{0}$, the algorithm is trying to find the shortest and collision-free path $\mu$. First, if there is a collision-free path between $\mathbf{n}_{bn}$ and the current node $\mathbf{n}_{0}$, the path $\mu$ is returned. If this is not the case, algorithm tries to find a collision-free path between some other node (denoted as $\mathbf{n}_{r}$) to the current node $\mathbf{n}_{0}$. When this is achieved, $\mathbf{n}_{r}$ becomes a new current node (denoted $\mathbf{n}_{shortest}$). We repeat this procedure until we can connect nodes $\mathbf{n}_{bn}$ and $\mathbf{n}_{0}$. 
When the robot returns to the best node, dead-end resolution process is considered finished and the standard exploration process continues.

\begin{figure}[t!]
	\centering
	\includegraphics[width=0.99\columnwidth]{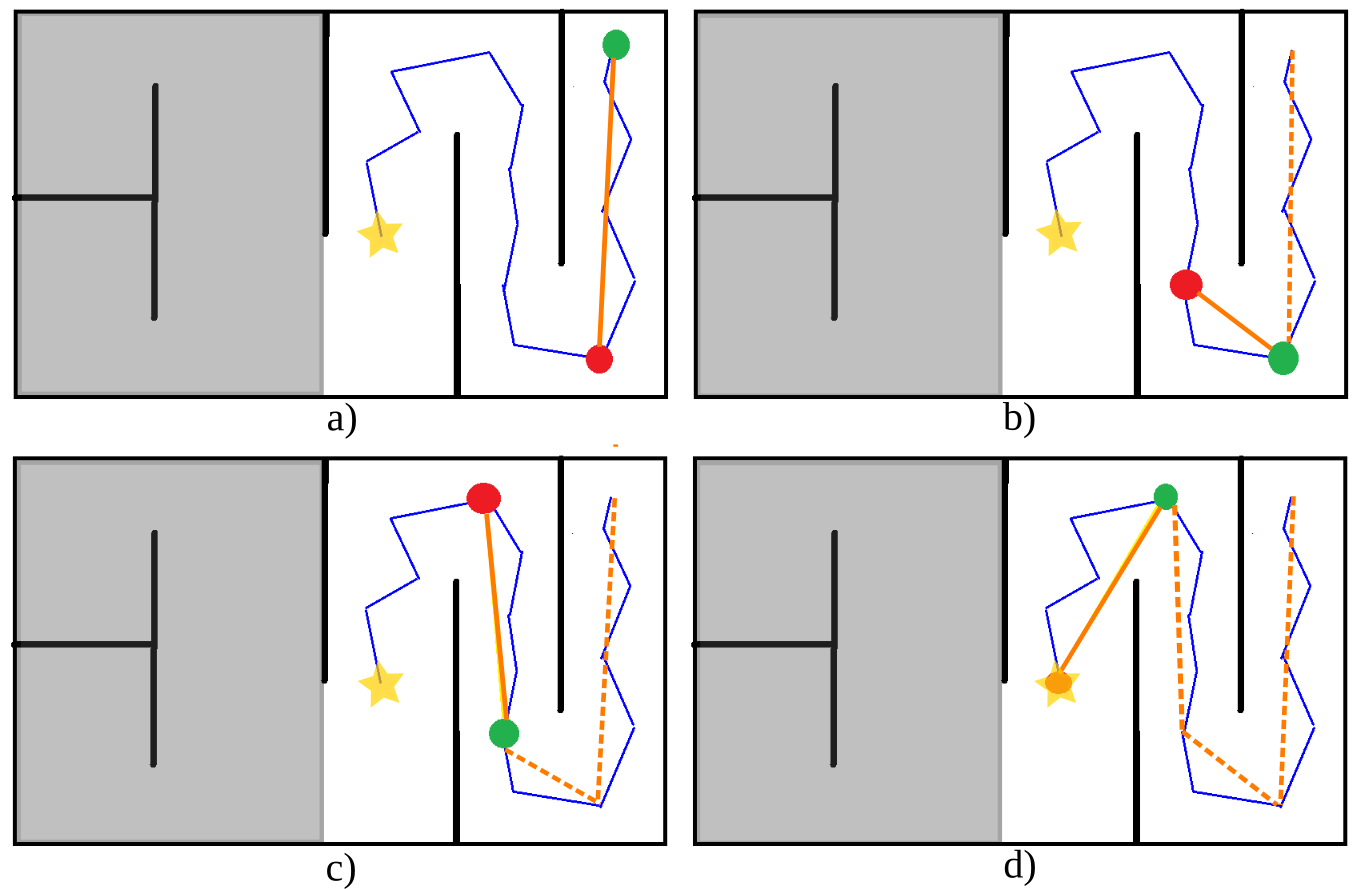}
	\caption{An illustration of returning from the dead end to the best node inside the proposed recovery strategy. The current node  $\mathbf{n}_{0}$ is marked green, while the yellow star denotes the best node $\mathbf{n}_{bn}$. The orange path connects the nodes from  $\mathbf{n}_{0}$ to $\mathbf{n}_{bn}$ to get the shortest possible collision-free path, which is then executed. Grey part of the illustration represents an unknown space. \vspace{-0.45cm}}
	\label{fig:history}
\end{figure}

Fig. \ref{fig:history} illustrates the process of dead end resolution by returning to the best node in a simple environment. The current node  $\mathbf{n}_{0}$ is marked green, while the yellow star denotes  $\mathbf{n}_{bn}$. The node $\mathbf{n}_{bn}$ is the one with the highest estimated information gain calculated using Eq. \ref{eq:node_gain}.   The algorithm calculates the shortest possible collision-free path (in orange color) up to $\mathbf{n}_{bn}$. Note that the simple, illustrative example in Fig. \ref{fig:history} might lead the reader to the conclusion that the robot can return following the blue path. Although this path will return the robot to the $\mathbf{n}_{bn}$, it would visit all previously planned RRT nodes. In complex environments, this can lead to significant and unnecessary visits to explored parts of the environment which then leads to the increase in return distance, time and energy consumption. The main purpose of this dead end resolving strategy is to shorten the return path and, therefore, avoid unnecessary visits to the previously explored environment.


\begin{algorithm}[t!]
\caption{Path returning to the best node}
\label{alg:recovery}
\begin{algorithmic}[1]

\STATE \textbf{Function:} 
\STATE \textbf{def} $\mu=getPath(\mathbf{n}_{first},\mathbf{n}_{last})$
\STATE $\mu = \emptyset$
\STATE $\mathbf{n}_{r} \leftarrow \mathbf{n}_{last}$
\WHILE{$\mathbf{n}_{r} \neq \mathbf{n}_{first}$}
\IF{Exist a collision-free path $\mu$ from $\mathbf{n}_{first}$ to $\mathbf{n}_{r}$}
\STATE Break
\ENDIF
\STATE $\mathbf{n}_{r} \leftarrow \mathbf{n}_{r-1}$
\ENDWHILE
\IF {$\mathbf{n}_{r} \neq \mathbf{n}_{last}$}
\STATE $\mu_0 \leftarrow$ $getPath(\mathbf{n}_{r},\mathbf{n}_{last})$
\STATE $\mu = \mu \cup \mu_0$
\ELSE \RETURN $\mathbf{n}_{last}$
\ENDIF
\RETURN $\mu$

\STATE
\STATE \textbf{Main:}
\STATE \textbf{Require} $[\mathbf{n}_{0}, \mathbf{n}_{1}, \dots , \mathbf{n}_{bn}]$ 
\STATE $\mu=getPath(\mathbf{n}_{0},\mathbf{n}_{bn})$

\end{algorithmic}
\end{algorithm}

\section{Simulation analysis}
\label{sec:simulation}

\subsection{System Setup}
Simulations are performed in the Gazebo environment using the Robot Operating System (ROS) and a model of the custom built \textit{Kopterworx} quadcopter. More details about our system and the control structure can be found in our previous work \cite{milijas2021}. The quadcopter is equipped with a Velodyne VLP-16 LiDAR sensor featuring a horizontal and vertical FOV $\alpha_{h} = 360^\circ$,  $\alpha_{v} = 30^\circ$, respectively. For collision checking, dimensions of a prism around the UAV are set to $l = 0.6$ m, $w = 0.6$ m , $h = 0.5$ m.
Parameters used in our experiments are shown in Table \ref{tab:parameters}. The proposed algorithm is compared with the RH-NBVP. Parameters $d^{planner}_{max} = 1.5$ m, $\lambda = 0.25$, $N_{max} = 20$ and the maximum RRT tree edge length of $1.5$ m refer to the setup of RH-NBVP explained in \cite{Bircher2016} and are set to indicated values. We adapted the NBVP to our quadcopter, equipped with a LiDAR and to our control system to allow the fairest possible comparison. 
We run three scenarios with different sizes and resolution $r$ and analyze the results. All simulations have been run 10 times on Intel(R)Core(TM) i7-10750H CPU @ 2.60GHz $\times$ 12.

\begin{table}[h!]
\centering
\caption{Exploration parameters}
\label{tab:parameters}
\begin{tabular}{l *4c}
\toprule
Parameter & Apartment & Maze & Large Maze \\
\midrule
$r$ [m] & 0.2, 0.4 & 0.1, 0.2 & 0.1, 0.2 \\
\midrule
$R_{max}$ [m] & 20.0 & 20.0 & 20.0 \\
\midrule
$\mathbf{v}^{\left\{x,y,z \right\}}_{max}$ [m/s] & 1.0 & 1.5 & 1.5 \\
\midrule
$\dot{\psi}_{max}$ [rad/s] & 0.8 & 0.8 & 0.8 \\
\midrule
$I_{range}$ [m] & 5.0 & 8.0 & 8.0 \\
\midrule
$\lambda$ & 0.3 & 0.6 & 0.6 \\
\bottomrule
\end{tabular}
\end{table}

\begin{table*}[!htbp]
\centering
\caption{Mean and standard deviation for the total exploration time $t_{exp}$ and the computational time per iteration $t_{c}$.}
\label{tab:results}
\begin{tabular}{l *5c}
\toprule
 &  & \multicolumn{2}{c}{\textbf{OURS}} & \multicolumn{2}{c}{\textbf{RH-NBVP}}\\
\midrule
\textbf{Scenario} & $\mathbf{r}$ [m] & $\mathbf{t_{c}}$ [ms] &  $\mathbf{t_{exp}}$[s] &  $\mathbf{t_{c}}$ [ms]&  $\mathbf{t_{exp}}$[s] \\
\midrule
\multirow{2}{*}{\textbf{Apartment}} & 0.4  & 4.41 $\pm$ 2.39 & 87.82$\pm$13.10 & 15.39$\pm$ 9.74 & 242.36$\pm$51.63 \\
    &0.2    &  19.63 $\pm$ 10.37 & 113.51$\pm$29.30 & 135.16 $\pm$ 57.86 & 276.84$\pm$70.54 \\
\midrule
\multirow{2}{*}{\textbf{Maze}} & 0.2 & 25.08 $\pm$ 10.89 & 209.24$\pm$31.03 & 383.33 $\pm$ 124.38 &504.566$\pm$75.23 \\
        & 0.1 &  81.61 $\pm$ 18.84 & 350.05$\pm$87.33  & 1024.19 $\pm$ 297.34 & 832.51$\pm$183.34 \\
\midrule
\multirow{2}{*}{\textbf{Large Maze}} & 0.2 & 48.67 $\pm$ 19.12 & 1017.23 $\pm$ 271.34 & 744.01 $\pm$ 244.53 & 1847.66 $\pm$ 305.78\\
        & 0.1 &  98.71 $\pm$ 37.52 & 1324.89 $\pm$ 283.22  & 2230.46 $\pm$ 579.43  & 2351.64 $\pm$ 547.52\\
\bottomrule
\end{tabular}
\end{table*}

The first scenario refers to a 10 m $\times$ 20 m $\times$ 3 m relatively simple apartment space used in \cite{Bircher2016}, \cite{Selin2019}, \cite{Dai2020}. The second scenario refers to a 20 m $\times$ 20 m $\times$ 2.5 m maze environment used in \cite{Oleynikova2017}, \cite{Dai2020}. Finally, the third simulation scenario refers to a 30 m $\times$ 30 m $\times$ 2 m large maze environment used in \cite{Witting2018}, \cite{Respall2021}. The robot performs a simple trajectory in a close proximity to the initial point, to ensure the planning is performed with some initial information. Additionally, we assume a reliable state estimation and focus on the exploration.

\subsection{Comparison of Raycasting and Shadowcasting}

We compared the performance of the RC (used in RH-NBVP) with our RSC-based planner in all three simulation scenarios and at different resolutions. The casting methods in the information gain calculation affect the computation time $t_c$ and thus the total exploration time. Computation times and total exploration times for all 10 runs are shown in Table \ref{tab:results}. It can be observed that the computation times for the RH-NBVP approach are significantly higher than in our approach, especially when using a high resolution map. Furthermore, the use of RC in RH-NBVP causes the computation time to increase as the complexity of the environment increases. When compared to the RH-NBVP, the computation time of our planner in the apartment scenario at the resolution $r = 0.2$ m is improved almost seven times. On the other hand, in the maze and large maze scenarios, our planner computation times are improved up to twenty times. The results have confirmed that the RC algorithm may cause a bottleneck in larger and more complex scenarios during the exploration. In other words, the robot has to stand still in the air for about 3 s to decide about the next best path. In simulation analysis, we noticed that setting the parameter inside which RC is performed,  $d^{planner}_{max}$,  to higher values leads higher computation times (up to 10 s).     
\subsection{Global Exploration Using Proposed Planner}
\begin{figure}[t!]
	\centering
	\includegraphics[width=0.98\columnwidth]{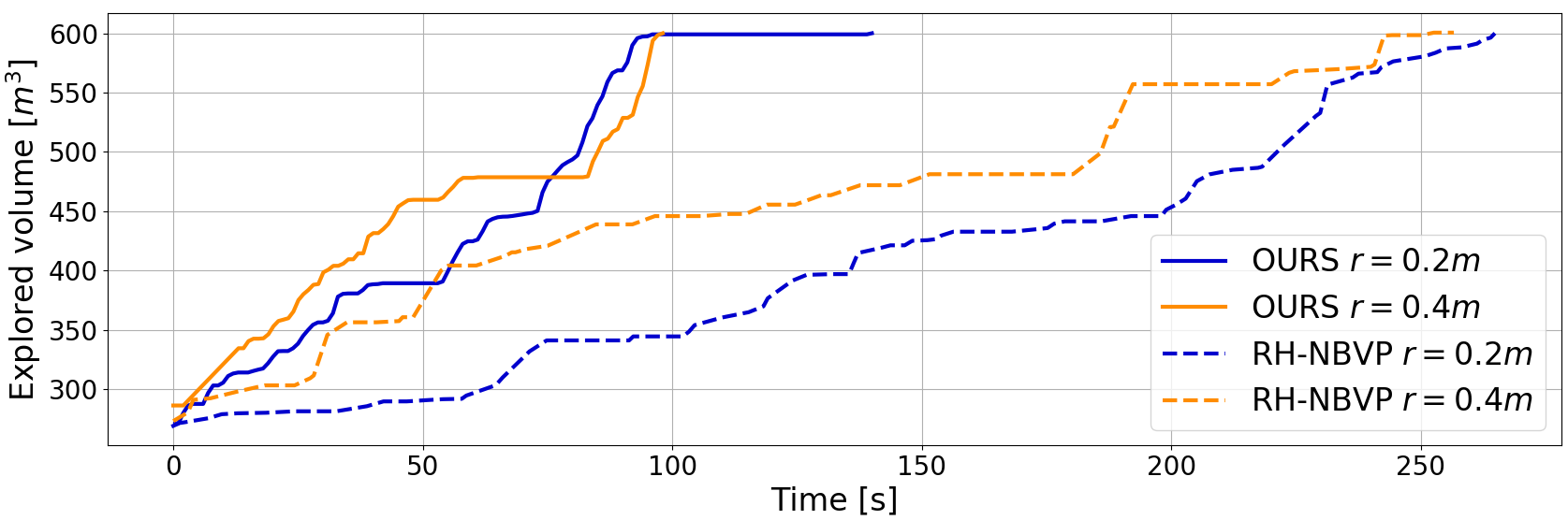}
	\caption{The explored volume in total exploration time for the apartment scenario. \vspace{-0.25cm}}
	\label{fig:apartment-volume}
\end{figure}

Several simulations were performed to compare the total exploration time of our exploration planner with the RH-NBVP. Fig. \ref{fig:apartment-volume} shows the explored volume over time for both algorithms at a voxel resolution of $r = 0.2$ m and $r= 0.4$ m. It can be observed that our planner explores the entire environment faster, especially when using a higher resolution. The graph shows that our method is significantly faster than RH-NBVP, taking less than 100 s to explore the apartment scenario at different map resolutions. 

\begin{figure}[t!]
	\centering
	\includegraphics[width=0.98\columnwidth]{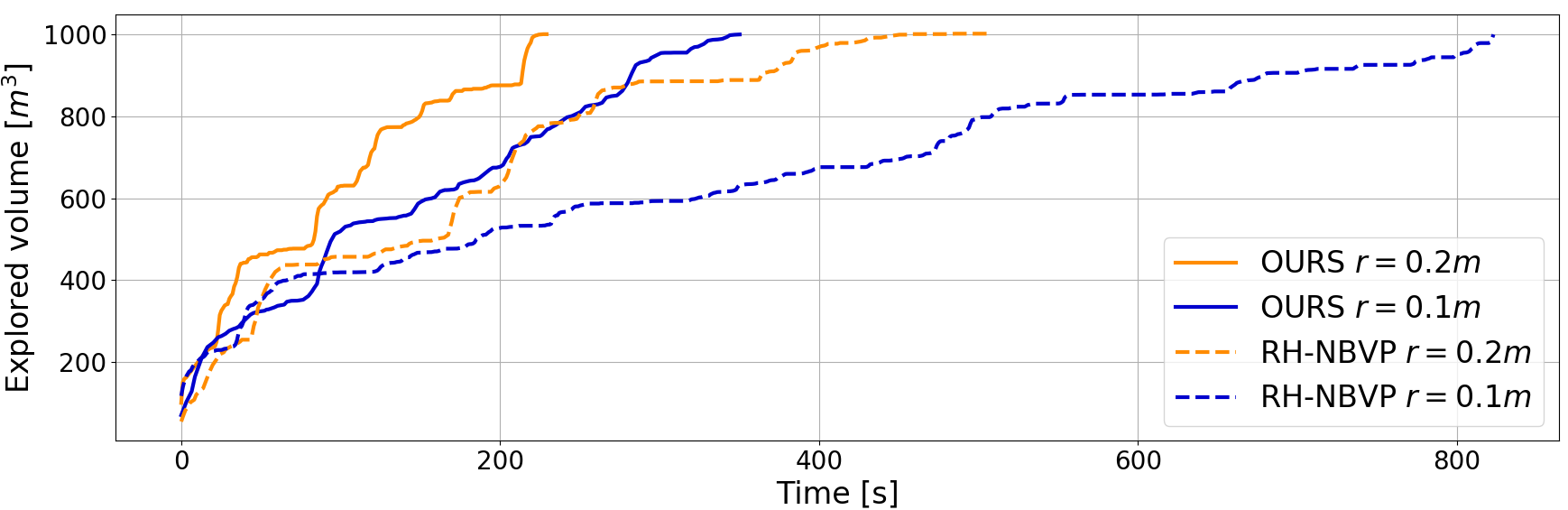}
	\caption{The explored volume in total exploration time for the maze scenario. \vspace{-0.25cm}}
	\label{fig:maze-volume}
\end{figure}

\begin{figure}[t!]
	\centering
	\includegraphics[width=0.65\columnwidth]{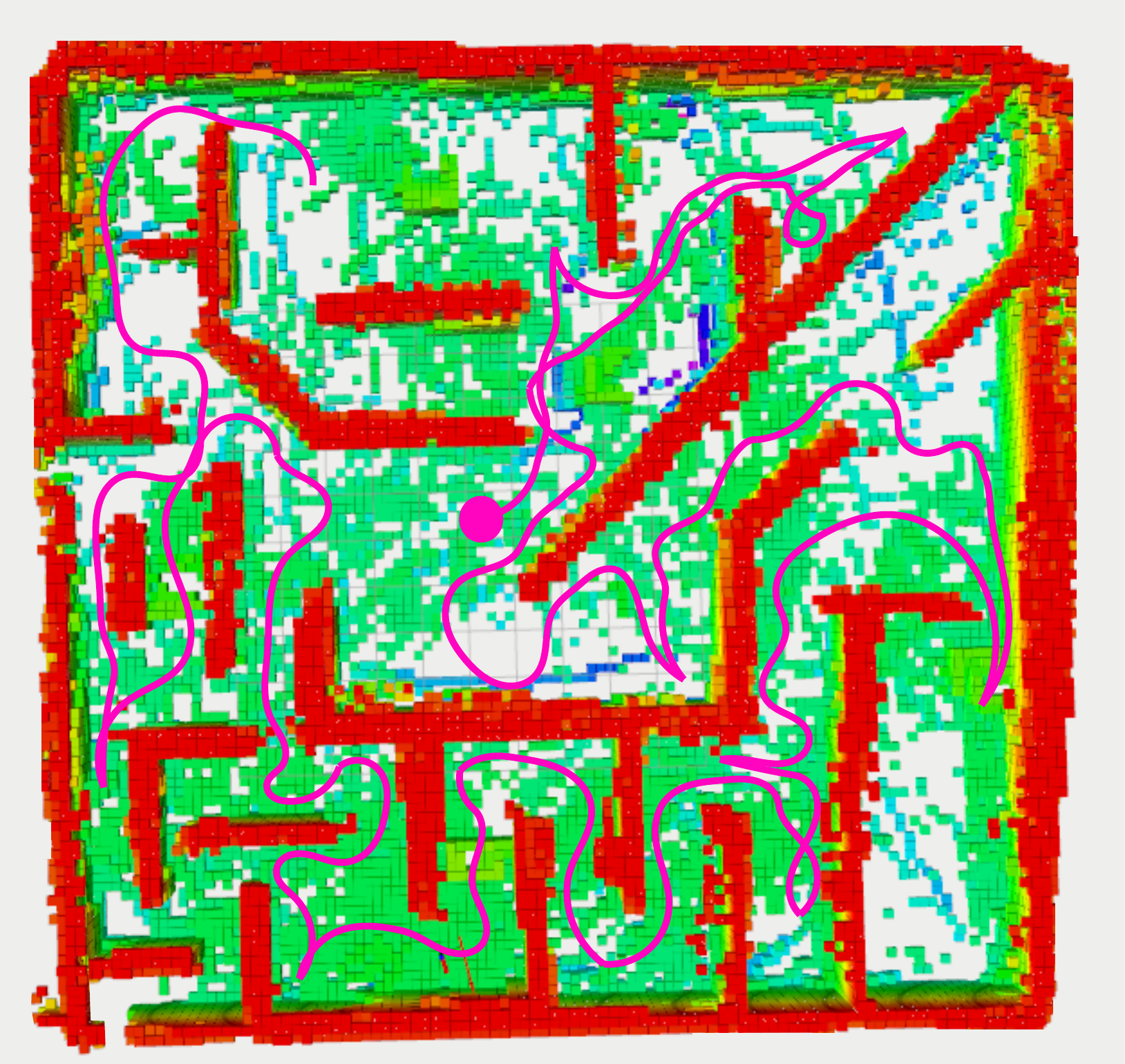}
	\caption{The OctoMap of the maze scenario generated by our exploration planner. The path traversed by the UAV during the exploration is marked in pink. The voxels are coloured according to their height. The starting position of the UAV is marked with a pink circle. \vspace{-0.25cm}}
	\label{fig:maze_path}
\end{figure}

In the maze scenario, both our algorithm and RH-NBVP were tested using a voxel resolution of $r = 0.1$ m and $r = 0.2$ m. We used higher resolutions because the environment contains some narrow corridors that the UAV cannot navigate through when a coarse resolution is used. The explored volume in time is shown in Fig. \ref{fig:maze-volume}. Our method explores the whole environment more than twice faster when compared to the RH-NBVP. Both planners behaves similarly at the beginning, but, as the time passes, the RH-NBVP shows its drawbacks, influencing the total exploration time. Our planner explores the maze environment in 209.24 $\pm$ 31.03 s and 350.05 $\pm$ 87.33 s for a resolution of 0.2 m and 0.1 m respectively. The total exploration time is comparable to the results from \cite{Dai2020}, which confirms the efficiency of our planner.
A thorough comparison of the experimental results with \cite{Dai2020} is not possible, due to different equipment and setup used, without the source code provided.
The OctoMap of the maze scenario generated by our planner at $r = 0.2$ m is shown in Fig. \ref{fig:maze_path} together with the corresponding UAV path.
Taking these results into consideration, it is shown that combining the cube-based approximation in the path information gain estimation, instead of considering the nodes only, results in a faster exploration. The random sampling of both our and the RH-NBVP algorithms leads to revisiting regions, but executing only the first node instead of the whole path results in a higher total exploration time for all three scenarios.

\subsection{Evaluation of the Dead End Resolving Strategy}
\begin{figure}[t!]
	\centering
	\includegraphics[width=0.98\columnwidth]{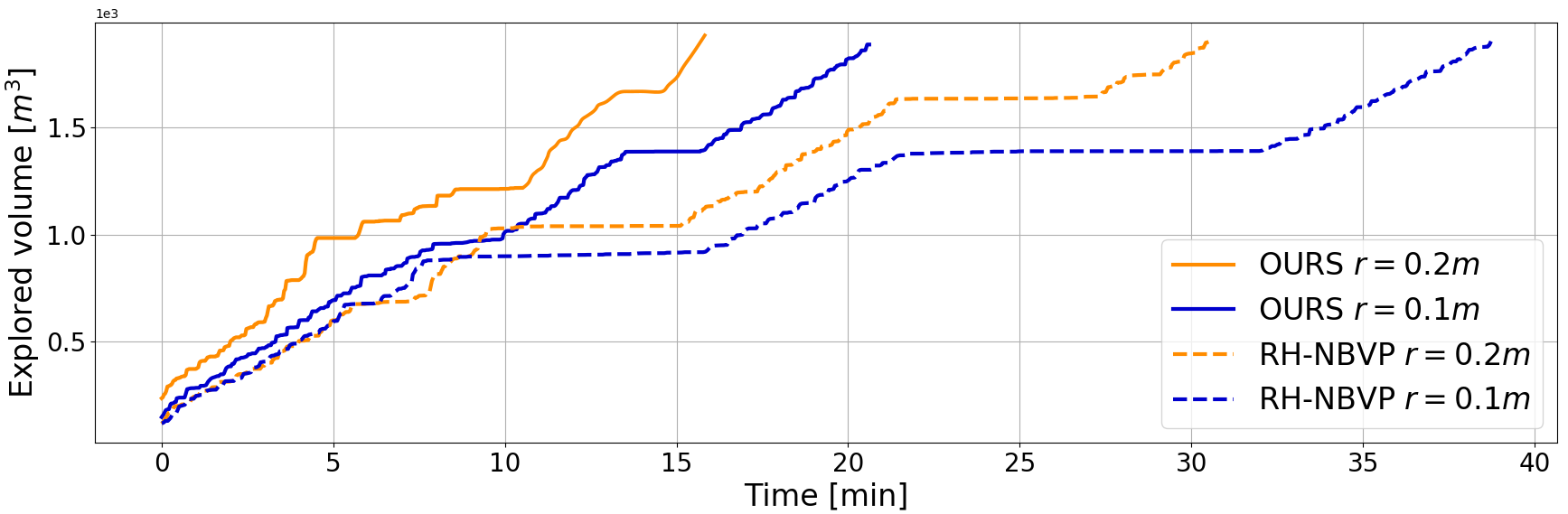}
	\caption{The explored volume in total exploration time for the large maze scenario. \vspace{-0.25cm}}
	\label{fig:large-maze-volume}
\end{figure}

The performance of the dead end resolving strategy is tested on a challenging large maze scenario with dead ends and narrow passages, and further compared to \cite{Bircher2016}. This experiment is to show the capabilities of the proposed method. The resolution is set to $r=0.1$ m and $r=0.2$ m for both planners. The results are shown in Fig. \ref{fig:large-maze-volume}. 
The results demonstrate that our algorithm completes the exploration in 17 minutes and 22 minutes for a resolution of 0.2 m and 0.1 m respectively.  That is more than two times faster on average than the NBVP and significantly more efficient. The graph shows that RH-NBVP spends large amount of time growing the tree when dead end states occur. When compared to other state-of-the-art results that use the large maze scenario and history tracking methods, \cite{Respall2021} reports time of 21 minutes at a resolution $r = 0.1$ m and $\mathbf{v}_{max} = 1.0 $ m/s, while \cite{Witting2018} finished the exploration in 30 minutes ($\mathbf{v}_{max} = 1.2$ m/s). Note that the system setup as well as the maximal exploration velocity are not the same as in our case.  

\begin{figure}[t!]
	\centering
	\includegraphics[width=0.99\columnwidth]{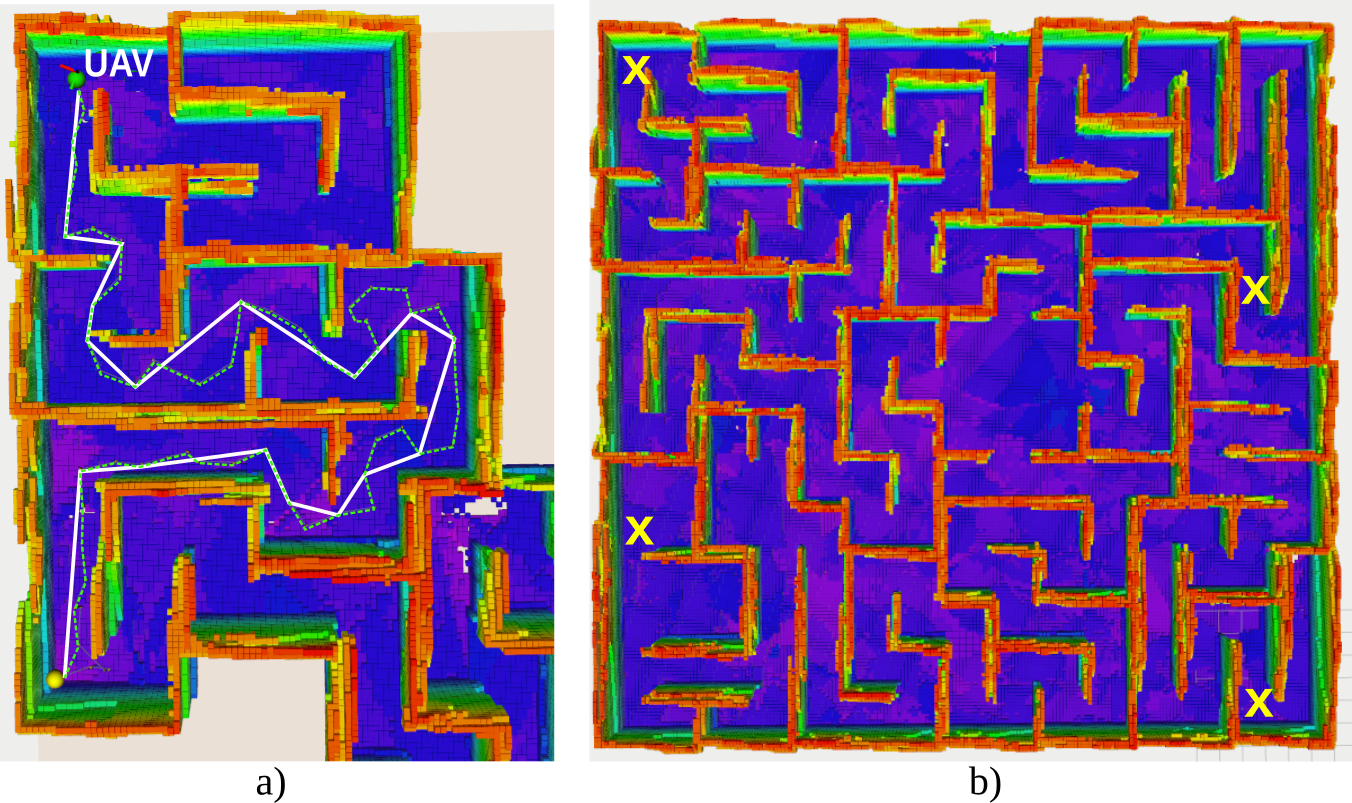}
	\caption{Large maze scenario. a) An instance of the strategy for resolving a dead end during the exploration. Position of the UAV is marked green, the goal position determined by the algorithm for resolving dead end is marked yellow, all nodes from the history list are connected with dashed lines while the executed path is marked white. b) An OctoMap created during exploration. Yellow markers represent positions of dead ends for specific runs. \vspace{-0.5cm}}
	\label{fig:lm-history}
\end{figure}

An instance of the proposed strategy for resolving dead ends is shown in Fig. \ref{fig:lm-history}. As can be observed, the algorithm successfully finds paths to the best node of all previously visited nodes and avoids growing a large RRT. Yellow markers in the Fig. \ref{fig:lm-history} b) represent positions of dead ends in a specific run. In such a large maze environment, multiple dead ends are expected and resolving them promptly leads to a more efficient exploration.

\section{Conclusion and future work}
\label{sec:conclusion}

This paper presents  a novel sampling-based planner for autonomous 3D exploration. The planner is capable of autonomously exploring a previously unknown bounded area and creating an OctoMap of the environment. The results showed an improved behaviour in terms of both computation and total exploration time compared to state-of-the-art strategies. The proposed information gain calculation and path evaluation ensures target evaluation in a short computation time, while a novel dead end recovery algorithm speeds up the exploration process.
This 3D exploration planner has been successfully tested and analysed in simulation scenarios and compared with state-of-the-are strategies.

For future work we consider testing our planner in an outdoor environment, as in our previous work \cite{Batinovic-RAL-2021} and extending our planner to a hybrid one, combined with the frontier-based approach. 
Video recordings of our exploration planner can be found at YouTube \cite{video}.

\section*{Acknowledgements}
\small{This work has been supported in part by the European Union through the European Regional Development Fund - The Competitiveness and Cohesion Operational Programme (KK.01.1.1.04.0041) through project named Heterogeneous autonomous robotic system in viticulture and mariculture (HEKTOR), and in part by EU-H2020 CSA project AeRoTwin - Twinning coordination action for spreading excellence in Aerial Robotics, grant agreement No. 810321. 
The work of doctoral student Ana Batinovic has been supported in part by the “Young researchers’ career development project--training of doctoral students” of the Croatian Science Foundation funded by the European Union from the European Social Fund.}

\bibliographystyle{ieeetr}
\typeout{}
\balance
\bibliography{Bibliography/exploration}

\end{document}